\documentclass[sigconf]{acmart}

\usepackage{amsmath}
\usepackage{booktabs}
\usepackage{tabularx}
\usepackage{adjustbox}
\usepackage{url}
\usepackage{mdwlist}
\usepackage{amsfonts}
\usepackage{multirow}
\usepackage{array}
\usepackage{epstopdf}
\usepackage{enumitem}
\usepackage[switch]{lineno}
\usepackage{subfigure}
\usepackage{bm}
\usepackage{dcolumn}

\newcommand{\mb}{\mathbf}

\newcommand{\mbb}{\mathbb}
\newcommand{\mc}{\mathcal}
\AtBeginDocument{%
  \providecommand\BibTeX{{%
    \normalfont B\kern-0.5em{\scshape i\kern-0.25em b}\kern-0.8em\TeX}}}

\copyrightyear{2024}
\acmYear{2024}
\setcopyright{acmlicensed}\acmConference[WWW '24 Companion]{Companion Proceedings of the ACM Web Conference 2024}{May 13--17, 2024}{Singapore, Singapore}
\acmBooktitle{Companion Proceedings of the ACM Web Conference 2024 (WWW '24 Companion), May 13--17, 2024, Singapore, Singapore}
\acmDOI{10.1145/3589335.3651560}
\acmISBN{979-8-4007-0172-6/24/05}

\begin{document}
\title{Simple Multigraph Convolution Networks}

\author{Danyang Wu}
\orcid{0000-0002-0309-1409}
\affiliation{%
  \institution{College of Information Engineering,\\Northwest A\&F University}
  \country{}
}
\author{Xinjie Shen}
\orcid{0009-0004-9176-5400}
\affiliation{%
  \institution{South China University of Technology}
    \country{}
}

\author{Jitao Lu}
\orcid{0000-0002-6065-2639}
\affiliation{%
  \institution{Northwestern Polytechnical University}
    \country{}
}
\author{Jin Xu}
\orcid{0009-0001-8735-3532}
\affiliation{%
  \institution{South China University of Technology\\ Pazhou Lab}
    \country{}
}
\author{Feiping Nie}
\orcid{0000-0002-0871-6519}
\affiliation{%
  \institution{Northwestern Polytechnical University}
    \country{}
}

\renewcommand{\shortauthors}{Wu, et al.}
\begin{abstract}
Existing multigraph convolution methods either ignore the cross-view interaction among multiple graphs, or induce extremely high computational cost due to standard cross-view polynomial operators. To alleviate this problem, this paper proposes a Simple MultiGraph Convolution Networks (SMGCN) which first extracts consistent cross-view topology from multigraphs including edge-level and subgraph-level topology, then performs polynomial expansion based on raw multigraphs and consistent topologies. In theory, SMGCN utilizes the consistent topologies in polynomial expansion rather than standard cross-view polynomial expansion, which performs credible cross-view spatial message-passing, follows the spectral convolution paradigm, and effectively reduces the complexity of standard polynomial expansion. In the simulations, experimental results demonstrate that SMGCN achieves state-of-the-art performance on ACM and DBLP multigraph benchmark datasets. Our codes are available at \href{https://github.com/frinkleko/SMGCN}{here}.
\end{abstract}

\maketitle
\section{Introduction}
\label{sec:intro}

Multigraph convolution is an important tool in graph machine learning and signal processing field, which aims to filter the graph signal in the spatial or (and) spectral domain of multiple graphs.

In recent years, multigraph convolution methods have been widely used in many fields, such as social network analysis~\cite{Sless2018,Cui2021}, recommendation system~\cite{Li2023,Zou2022}, and bioinformatics~\cite{Yan2019,Bhadra2019}, for their data is usually represented as a multigraph, which contains multiple different views' graphs representing different relation about the same object. For example, in social network analysis, node can be user, and different views can be different social relations between users, such as friendship, following, and co-occurrence.

In recent years, several multigraph convolution methods have been proposed, including building robust node features to help to regularize and aggregate embeddings learned from each view~\cite{Qu2017}, parallel learning embeddings of each view and concatenate their min, max and mean as final embedding ~\cite{pgcn}, summing embeddings learned from different orders of each view's graphs ~\cite{Butler2023} and Multi-Input Multi-Output GCN (MIMO-GCN)~\cite{Butler2023} that computes embeddings learned from all combination of different views' graphs.

\begin{figure}[t]
  \centering
  \includegraphics[width=0.9\linewidth]{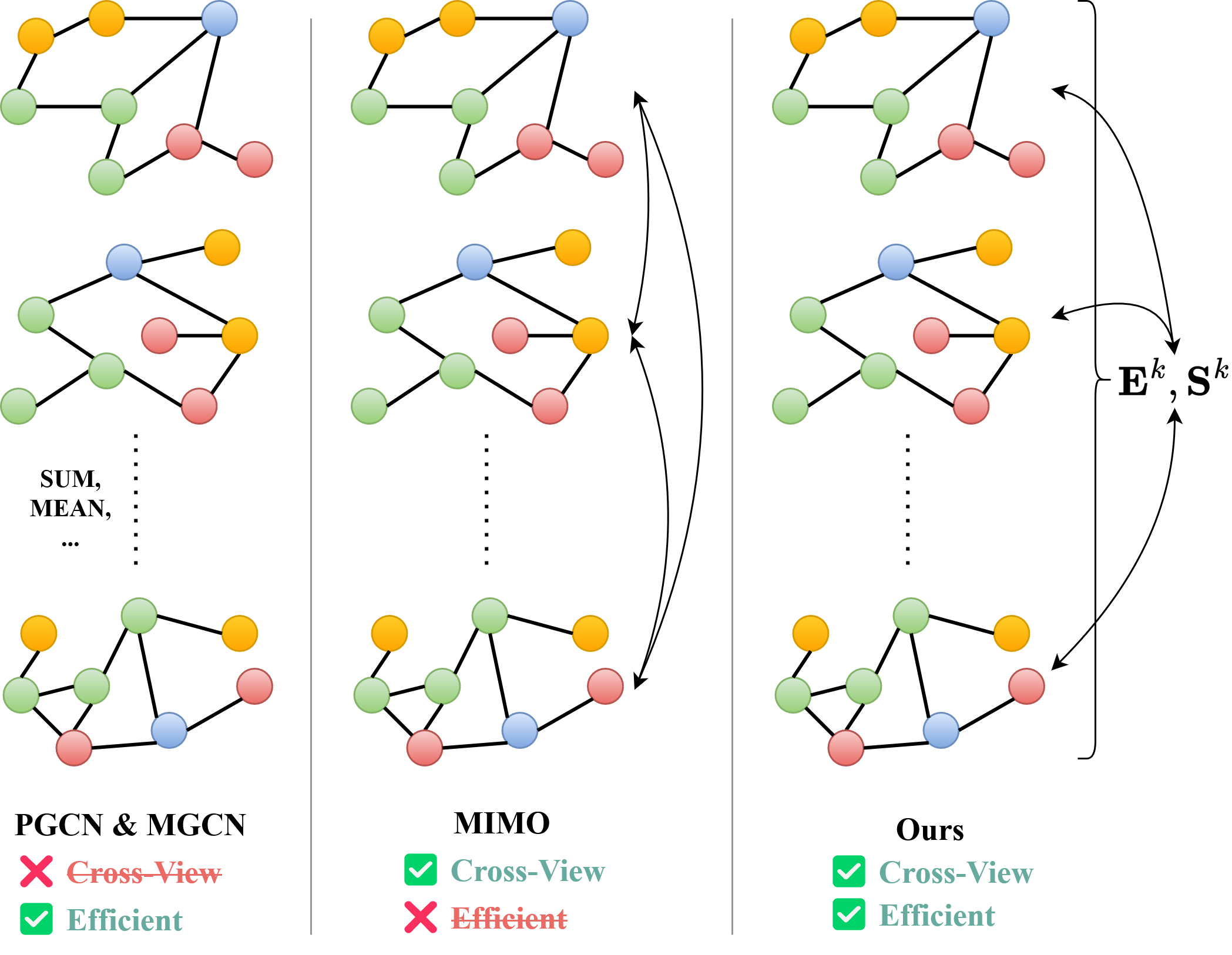}
    \vspace{-0.4cm}
  \caption{Overview of the proposed SMGCN.}
  \label{fig:overview}
    \vspace{-0.75cm}
  
\end{figure}

However, existing multigraph convolution methods still have difficult in effectively solving the conflict between effectiveness and efficiency. As Figure~\ref{fig:overview} shows, on the one hand, a part of methods consider integrate the multigraphs directly such as selection or weighted sum operator, which is efficient but ignores the cross-view interaction among multiple graphs. On the other hands, several methods such as MIMO~\cite{Butler2023}, performs standard cross-view polynomial expansions on multigraphs, which integrates the high-order structure with a global perspective, but the computational cost can rapidly increase based on the number of polynomial orders and views. To alleviate this conflict, this paper proposes a Simple MultiGraph Convolution Networks (SMGCN) method which contains consistent topology extraction module and simple polynomial expansion module. To be specific, SMGCN first considers extracting the consistent cross-view topology in edge-level and subgraph-level, then performs polynomial expansion between raw multigraphs edge-level, subgraph-level to induce the filtered graph signal. More importantly, SMGCN is a simple and efficient method, which also performs credible cross-view spatial message-passing and follows the spectral convolution paradigm. In experiments, we evaluate SMGCN on two famous multigraph benchmark datasets, including ACM and DBLP. The experiment results demonstrate the superiorities of SMGCN with the perspective of effectiveness and efficiency.

The main contributions of this paper are summarized into the following two folds:
\begin{itemize}
  \item[•] In theory, SMGCN performs cross-view interaction via a simple and efficient spectral convolution focusing on extracting edge-level and subgraph-level consistent topology. It effectively alleviates the conflict between effectiveness and efficiency in previous works.
  \item[•] In the simulations, experiment results demonstrate that SMGCN can achieve state-of-the-art performance with lower computational cost.
\end{itemize}

\textbf{Notations.}
In this paper, a $m$ views multigraph is denoted as $\mc{G} = (\mc{V},\{\mc{E}^{(1)} \ldots \mc{E}^{(m)} \})$\footnote{We treat $m$ default bigger than 2 for standard cross-view polynomial expansion can be used directly when $m=2$.}. $\mc{V}$ is the set of nodes, $\mc{E}^{(v)}$ is the set of edges in view $v$. For easily understanding, we use $\{ \mb{A}^{(1)},\ldots,\mb{A}^{(m)} \}$ to represent the adjacency matrix of each view's graph, where $\mb{A}^{(v)} \in \mbb{R}^{n \times n}$ is the adjacency matrix of view $v$ and $n$ is the number of nodes. $\mb{D}^{(m)}$ is the degree matrix of view $v$, where $\mb{D}^{(v)}_{i,i} = \sum_{j=1}^n \mb{A}^{(v)}_{i,j}$. $\mb{X} \in \mbb{R}^{n \times d}$ is the node feature matrix, where $d$ is the dimension of node feature. $\mb{I}$ is the identity matrix. $\mb{W}$ is the learnable projection matrix. $K$ is the order of polynomial expansion.

\section{Related Works}
In this section, we introduce some representative works about multigrpah convolution methods, including Parallel GCN (P-GCN)~\cite{pgcn}, Merged GCN (M-GCN)~\cite{Butler2023} and Multi-Input Multi-Output GCN (MIMO-GCN)~\cite{Butler2023}.

\textbf{1) P-GCN} is the fundamental method to perform multigraph convolution calculation, which aims to learn node embedding based on each view's graph, then concatenate the min, max and mean of these embeddings as final embedding. This method only focus on the first-order information of each view's graph separately. The concrete formulation can be written as follows:
\begin{equation}
  \label{eq:pgcn}
  \begin{aligned}
     & \mb{Z}^{(v)} = \mb{D}^{(v)-\frac{1}{2}}\mb{A}^{(v)}\mb{D}^{(v)-\frac{1}{2}}\mb{X}\mb{W}^{(v)} \quad \forall v \in \{1,\ldots,m\}, \\
     & \mb{Z}= \mb{Concat}(\mb{min}(\{\mb{Z}^{(1)},\ldots,\mb{Z}^{(m)}\} ),                                                              \\
     & \quad \quad \mb{max}(\{\mb{Z}^{(1)},\ldots,\mb{Z}^{(m)}\} ),\mb{mean}(\{\mb{Z}^{(1)},\ldots,\mb{Z}^{(m)}\} ),
  \end{aligned}
\end{equation}
where $\mb{W}^{(v)}$ is the projection matrix for graph in view $v$ and $\mb{Z}^{(v)}$ is the node embedding matrix for graph in view $v$. $\mb{Z} \in \mbb{R}^{n \times c}$ is the final node embedding matrix, where $c$ is the dimension of node embedding.
$\mb{min}(\cdot)$, $\mb{max}(\cdot)$ and $\mb{mean}(\cdot)$ are set operators, which compute minimum, maximum and mean values across all elements in all inputs, respectively.

\textbf{2) M-GCN} is the fundamental method to perform multigraph convolution calculation, which considers the high-order information of each view's graph, then summing the embeddings learned from different orders of each view's graphs as final embedding. This method focus on different orders of each view's graph separately. The concrete formulation can be written as follows:
\begin{equation}
  \mb{Z} = \mb{I}\mb{X}\mb{W}_{\mb{I}} + \sum_{i=1}^{m}\sum_{k=1}^{K}(\mb{A}^{(i)})^k\mb{X}\mb{W}_{(\mb{A}^{(i)})^k},
\end{equation}
where $\mb{W}_{\mb{I}}$ is the projection matrix for identity matrix, $\mb{W}_{(\mb{A}^{(i)})^k}$ is the projection matrix for $k$-th order of $\mb{A}^{(i)}$.

\textbf{3) MIMO-GCN} is the state-of-the-art method to perform multigraph convolution calculation, which aims to among high-order interaction of multiple graphs, considers any arrangement of the views' graphs with given order $K$. Then embeddings learned from all combination of different views' graphs are summed as final embedding. This method focus on interaction among different views' graph of different orders. The concrete formulation can be written as follows:
\begin{equation}
  \mb{Z} = \mb{I}\mb{X}\mb{W}_{\mb{I}} + \mb{Poly}(\{\mb{A}^{(1)},\ldots,\mb{A}^{(m)}\},K),
\end{equation}
where $\mb{Poly}(\{\mb{A}^{(1)},\ldots,\mb{A}^{(m)}\},K)$ is the polynomial expansion of $\{\mb{A}^{(1)},\ldots,\mb{A}^{(m)}\}$ with order $K$. It can be written as follows:
\begin{equation}
  \mb{Poly}(\{\mb{A}^{(1)},\ldots,\mb{A}^{(m)}\},K) = \sum_{k=1}^K \sum_{p_k \in P_k} p_k \mb{X}\mb{W}_{p_k},
\end{equation}
where $P_k$ is the set of all possible arrangements of length $k$ of $\{\mb{A}^{(1)},\ldots,\mb{A}^{(m)}\}$, $\mb{W}_{p_k}$ is the projection matrix for $p_k$.

\section{Simple Multigraph Convolution Networks (SMGCN)}
In this section, we first propose the edge-level and subgraph-level consistent topology extraction modules, then introduce the proposed SMGCN model.

\subsection{Credible Edge-level Topology Extraction} Given multigraph $\{\mb{A}^{(1)},\ldots \mb{A}^{(m)}\}$, we first calculate the sum of all views $\sum_{v=1}^m \mb{A}^{(v)}$, then we filter the raw edge via cross-view voting scheme, $i.e.$, if an edge exists in two views, we set it as a credible raw edge. The concrete formulation can be written as follows:
\begin{align} \label{edge-level}
  e_{i,j} =
  \begin{cases}
    1, & \text{if}~ \sum_{v=1}^m \mc{I}(a^{(v)}_{i,j}>0) \geqslant 2, \\
    0, & \text{Otherwise},
  \end{cases}
\end{align}
where $\mb{E} \in \mbb{R}^{n \times n}$ is the similarity matrix represented the edge-level credible topology. Obviously that since $\mb{E}$ can hold part of edges based on cross-view topology, it can be seen as a summarization of raw multigraph $\{\mb{A}^{(1)},\ldots \mb{A}^{(m)}\}$, thus it can be used as a representative structure to perform high-order interaction with raw multigraph.
\subsection{Credible Subgraph-level Topology Extraction}
The exacted edge-level topology aforementioned above summarizes the credible topology structures of the raw multigraphs, which focuses on the global structure of the graph of each view. In this subsection, we focus on extracting credible subgraph-level topology, $i.e.$, latent subgraph structure. Formally, \footnote{If $\{\mb{A}^{(1)},\ldots \mb{A}^{(m)}\}$ are weighted graph, this step can be skipped.}suppose $\{\mb{A}^{(1)},\ldots \mb{A}^{(m)}\}$ are unweighted graph, we first calculate the triangle-based weighted similarity matrix for each view as follows:
\begin{align} \label{subgraph-level-selftri}
  \mb{\widehat{A}}^{(v)} = \left(\mb{A}^{(v)}+\mb{I}\right)^2 \circ \left(\mb{A}^{(v)}+\mb{I}\right).
\end{align}
\begin{figure}[t]
  \centering
  \includegraphics[width=.7\linewidth]{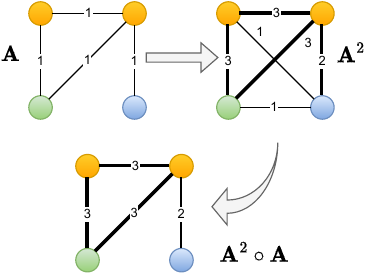}
  
  \caption{Triangle-based weighted similarity matrix generation. Self-loop is omitted in this figure.}
  \vspace{-0.5cm}
  \label{fig:1nn}
\end{figure}
In Figure~\ref{fig:1nn}, we illustrate the triangle-based weighted similarity matrix generation procedure.
Then we extract the first nearest neighbor matrix $\widehat{\mb{A}}^{(v)}$ for each view as follows:
\begin{align} \label{subgraph-level-1nn}
  \widehat{a}^{(v)}_{i,j} =
  \begin{cases}
    1, & \text{if}~ i \in \mc{N}_1^{(v)}(j) ~\text{or}~ j \in \mc{N}_1^{(v)}(i), \\
    0, & \text{Otherwise},
  \end{cases}
\end{align}
where $\mc{N}_1^{(v)}(i)$ is set of the nearest neighbor of sample $i$ in $v$-th view. Afterwards, according to the same principle of credible edge-level topology extraction, we calculate the sum of all views $\sum_{v=1}^m \widehat{\mb{A}}^{(v)}$, then we filter the edge via cross-view voting scheme as follows:
\begin{align} \label{edge-level-S}
  s_{i,j} =
  \begin{cases}
    1, & \text{if}~ \sum_{v=1}^m \mc{I}(\widehat{a}^{(v)}_{i,j}>0) \geqslant 2, \\
    0, & \text{Otherwise},
  \end{cases}
\end{align}
where $\mb{S} \in \mbb{R}^{n \times n}$ is the similarity matrix represented the edge-level credible topology. Naturally, $\mb{S}$ can be seen as a summarization of generated subgraph-level multigraph similarity matrices $\{\tilde{\mb{A}}^{(1)},\ldots \tilde{\mb{A}}^{(m)}\}$. More importantly, $\mb{S}$ may generate the new edges from raw multigraph due to the sub-graph generation procedure, which provides new perspective in the cross-view interactions of multigraph.
\subsection{Simple High-order Interaction Formulation}
According to the credible edge-level topology $\mb{E}$ and subgraph-level topology $\mb{S}$, we design the following simple high-order interaction formulation (For better illustration, we denote $\mb{T}^{(1)} = \mb{E}$ and $\mb{T}^{(2)} = \mb{S}$) to obtain the hidden layer multigraph representation\footnote{$(\mathbf{A}^{(v)})^k (\mathbf{T}^{(t)})^{K-k}$ can be symmetrized in practice (optional).}:
\begin{align} \label{SMGCN}
  \mathbf{Z} = \mathbf{I}\mathbf{X}\mathbf{W_I} +  \sum_{t=1}^2\sum_{v=1}^{m}\sum_{k=1}^{K} (\mathbf{A}^{(v)})^k (\mathbf{T}^{(t)})^{K-k}\mathbf{X}\mathbf{W}_{t,v,k},
\end{align}
where $\mathbf{W}_{\mb{I}}$ is the projection matrix for node feature $\mb{I}\mb{X}$ and $\mathbf{W}_{t,v,k}$ is the projection matrix for each polynomial combination for $\{\mb{A}^{(1)},\ldots,$ $\mb{A}^{(m)}\}$ and $\{\mb{T}^{(1)},\mb{T}^{(2)}\}$, $K$ is the order of polynomial expansion. Compared with previous works such as the standard cross-view polynomial expansion of MIMO, SMGCN cleverly performs the cross-view interaction of multigraph via extracting two credible cross-view topology $\{\mb{T}^{(1)},\mb{T}^{(2)}\}$, which significantly reduces the computational cost. More importantly, SMGCN can be seen as a pruning form of the standard cross-view high-order polynomial interaction among $\{\mb{A}^{(1)},\ldots,\mb{A}^{(m)},\mb{T}^{(1)},\mb{T}^{(2)}\}$, thus it also satisfies the spectral properties of standard polynomial expansion, such as Theorem 1 and Lemma 2 in ~\cite{Butler2023}.

\textbf{Computational Cost Comparison.} We can evaluate the computational cost of SMGCN based on the number of parameters. For instance, suppose the size of all projection matrices (including $\mb{W_I}$ and $\mb{W}_{t,v,k}$) is $d_0 \times d_1$ and the order is $K$. In SMGCN, the number of parameters is $O(mKd_0d_1)$, which is linear $w.r.t$ the polynomial order. However, due to the standard cross-view polynomial expansion, the number of parameters of MIMO (without the post-processing pruning operation) is $A_{m+K-1}^{m-1} = \frac{m+K-1!}{m-1!}$. According to Lemma 1 in ~\cite{Butler2023}, the post-processing pruning operation can reduce a part of terms when $K>3$, but the concrete computational cost is hard to evaluate since it is determined by threshold.

\section{Experiments}
\subsection{Benchmark Datasets}
In this section, we introduce the benchmark datasets used in our experiments, including ACM~\cite{ACMDataset}, DBLP~\cite{DBLPDataset}. To perform task on multi-view graph, we follow the usage in ~\cite{Fan2020}, which turns graph with multi relational edges and multi types nodes into multi-view graphs with only one type nodes. The statistics of these datasets are summarized in Table~\ref{tab:dataset}.

For ACM dataset, we consider four views, including co-author, co-subject, co-cite and co-refer relation between papers. For DBLP dataset, we consider three views, including co-paper, co-conference and co-term relation between authors. All the datasets are available at Pytorch Geometric~\cite{pyg}.

\begin{table}[!t]
  \centering
  \begin{adjustbox}{width=\columnwidth,center}
    \begin{tabular}{|c|c|c|c|c|c|}
      \hline
      Dataset                                                      &
      \begin{tabular}[c]{@{}c@{}}Target\\ Node\\ Type\end{tabular} &
      \begin{tabular}[c]{@{}c@{}}Node\\ Number\end{tabular}        &
      \begin{tabular}[c]{@{}c@{}}Node\\ Feature\end{tabular}       &
      Views                                                        &
      \begin{tabular}[c]{@{}c@{}}View Names and Edge Number\end{tabular}                                                     \\ \hline
      ACM                                                          &
      Paper                                                        &
      3025                                                         &
      1902                                                         &
      4                                                            &
      \begin{tabular}[c]{@{}c@{}}Co-author(30752),\\  Co-subject(2214064),\\  Co-cite(5343),\\  Co-refer(5343).\end{tabular} \\ \hline
      DBLP                                                         &
      Author                                                       &
      4057                                                         &
      334                                                          &
      3                                                            &
      \begin{tabular}[c]{@{}c@{}}Co-paper(6572),\\ Co-conference(3225446),\\ Co-term(7611879).\end{tabular}                  \\ \hline
    \end{tabular}
  \end{adjustbox}
  
  \caption{Statistics of benchmark datasets.}
  \label{tab:dataset}
  \vspace{-0.9cm}
\end{table}

\begin{table}[!t]
  \centering
  \begin{adjustbox}{width=\columnwidth,center}
    \begin{tabular}{|cl|ccc|ccc|}
      \hline
      \multicolumn{2}{|c|}{\multirow{2}{*}{Model}} &
      \multicolumn{3}{c|}{ACM(4 views)}            &
      \multicolumn{3}{c|}{DBLP(3 views)}                                                                                                                                 \\ \cline{3-8}
      \multicolumn{2}{|c|}{}                       & \multicolumn{1}{c|}{ACC} & \multicolumn{1}{c|}{F1} & NMI & \multicolumn{1}{c|}{ACC} & \multicolumn{1}{c|}{F1} & NMI \\ \hline
      \multicolumn{2}{|c|}{P-GCN}                  &
      \multicolumn{1}{c|}{93.77}                   &
      \multicolumn{1}{c|}{93.82}                   &
      76.89                                        &
      \multicolumn{1}{c|}{91.43}                   &
      \multicolumn{1}{c|}{90.68}                   &
      73.31                                                                                                                                                              \\ \hline
      \multicolumn{2}{|c|}{M-GCN}                  &
      \multicolumn{1}{c|}{94.10}                   &
      \multicolumn{1}{c|}{94.15}                   &
      77.94                                        &
      \multicolumn{1}{c|}{93.12}                   &
      \multicolumn{1}{c|}{92.47}                   &
      77.48                                                                                                                                                              \\ \hline
      \multicolumn{2}{|c|}{MIMO-GCN}               &
      \multicolumn{1}{c|}{94.33}                   &
      \multicolumn{1}{c|}{94.40}                   &
      79.04                                        &
      \multicolumn{1}{c|}{92.57}                   &
      \multicolumn{1}{c|}{91.86}                   &
      76.04                                                                                                                                                              \\ \hline
      \multicolumn{2}{|c|}{SMGCN}                  &
      \multicolumn{1}{c|}{\textbf{95.04}}          &
      \multicolumn{1}{c|}{\textbf{95.11}}          &
      \textbf{80.80}                               &
      \multicolumn{1}{c|}{\textbf{93.40}}          &
      \multicolumn{1}{c|}{\textbf{92.78}}          &
      \textbf{78.15}                                                                                                                                                     \\ \hline
    \end{tabular}
  \end{adjustbox}
  
  \caption{Metrics of different methods on ACM and DBLP dataset.}
  \label{tab:results}
  \vspace{-1cm}
  
\end{table}

\begin{figure}[!t]
  \centering
  \begin{minipage}[h]{\linewidth}
    \centering
    \centerline{\includegraphics[width=\linewidth]{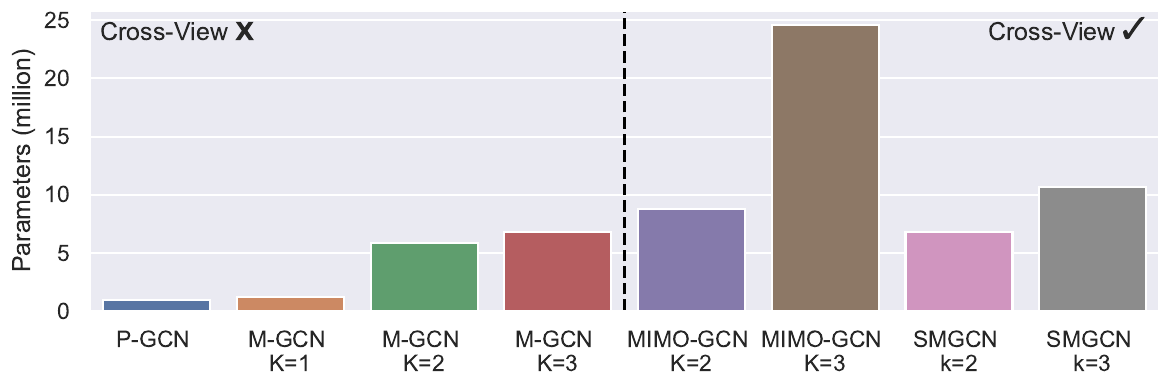}}
    \centerline{(a) Parameters of models on ACM dataset}\medskip
  \end{minipage}
  \begin{minipage}[h]{\linewidth}
    \centering
    \centerline{\includegraphics[width=\linewidth]{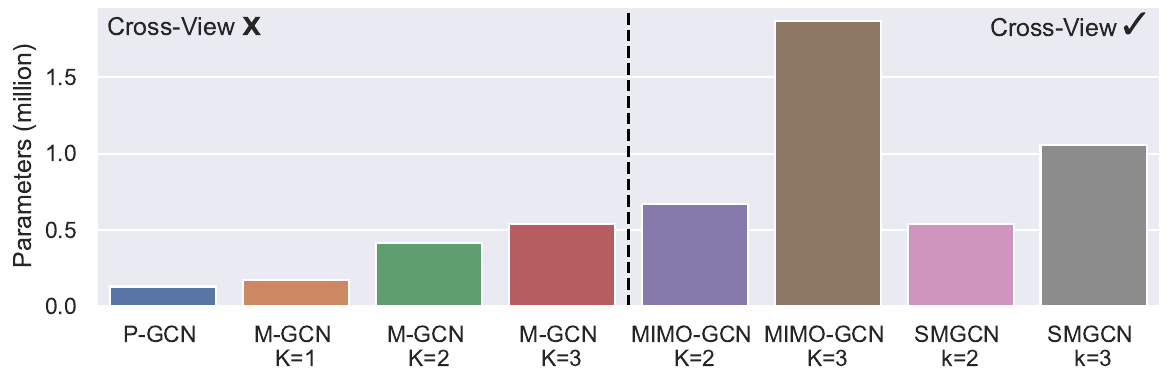}}
    \centerline{(b) Parameters of models on BDLP dataset}\medskip
  \end{minipage}
    \vspace{-0.5cm}
  \caption{Parameters of models on ACM and DBLP dataset.}
  \label{fig:parameters}
  
\end{figure}
\subsection{Experimental Settings and Evaluation Metrics}
We perform node classification tasks on both datasets, with given train/test split in corresponding dataset. 10\% of nodes in original training set are used as validation set. We grid search the learning rate and weight decay in $\{$0.1,0.01,0.001$\}$ and $\{$0,1e-5.1e-4,1e-3,1e-2$\}$. All the optimizer are Adam. All the hidden dimension are set to 128. All the activation function are ReLu. One linear layer is used to project the node embedding to the classification space. Due to memory limitation, we set the maximum order of polynomial expansion $K$ to 3. Accuracy, F1 macro and NMI are used as evaluation metrics.

\subsection{Results and Analysis}
Results are shown in Table~\ref{tab:results}. We can see that our method outperforms all the baseline methods on both datasets, achieving state-of-the-art performance.

\begin{figure}[!t]
    \vspace{-0.5cm}
  \centering
  \begin{minipage}[h]{0.45\linewidth}
    \centering
    \centerline{\includegraphics[width=\linewidth]{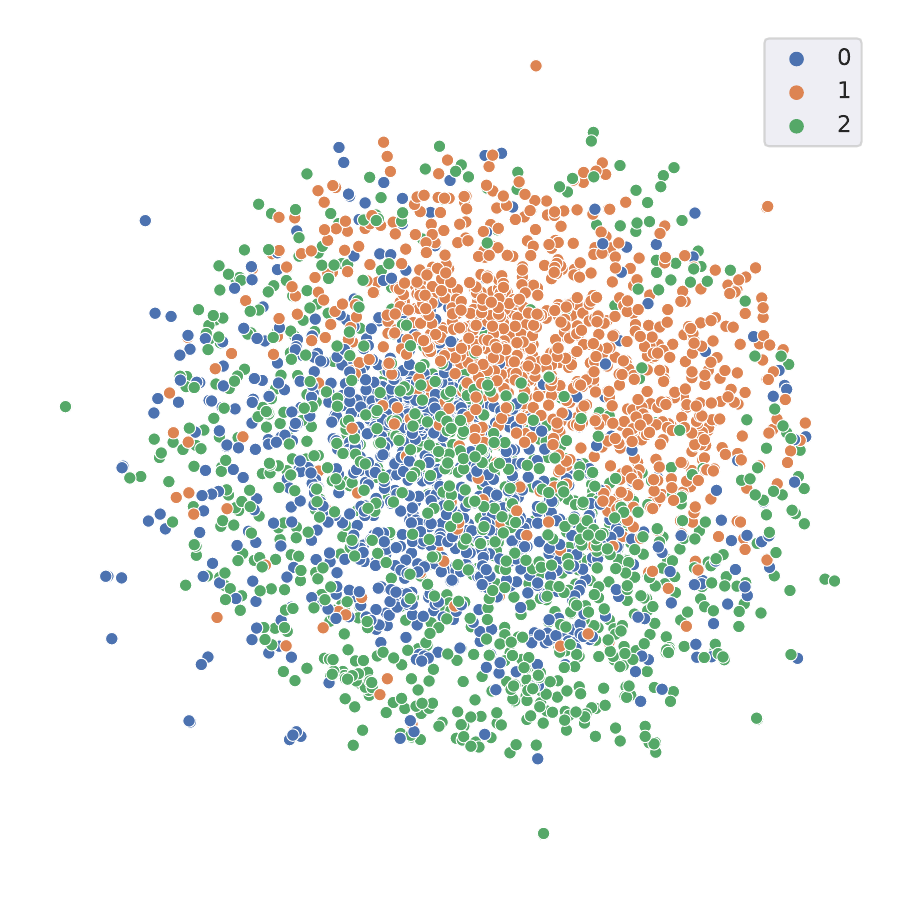}}
    \centerline{(a) Node features}\medskip
  \end{minipage}
  \begin{minipage}[h]{0.45\linewidth}
    \centering
    \centerline{\includegraphics[width=\linewidth]{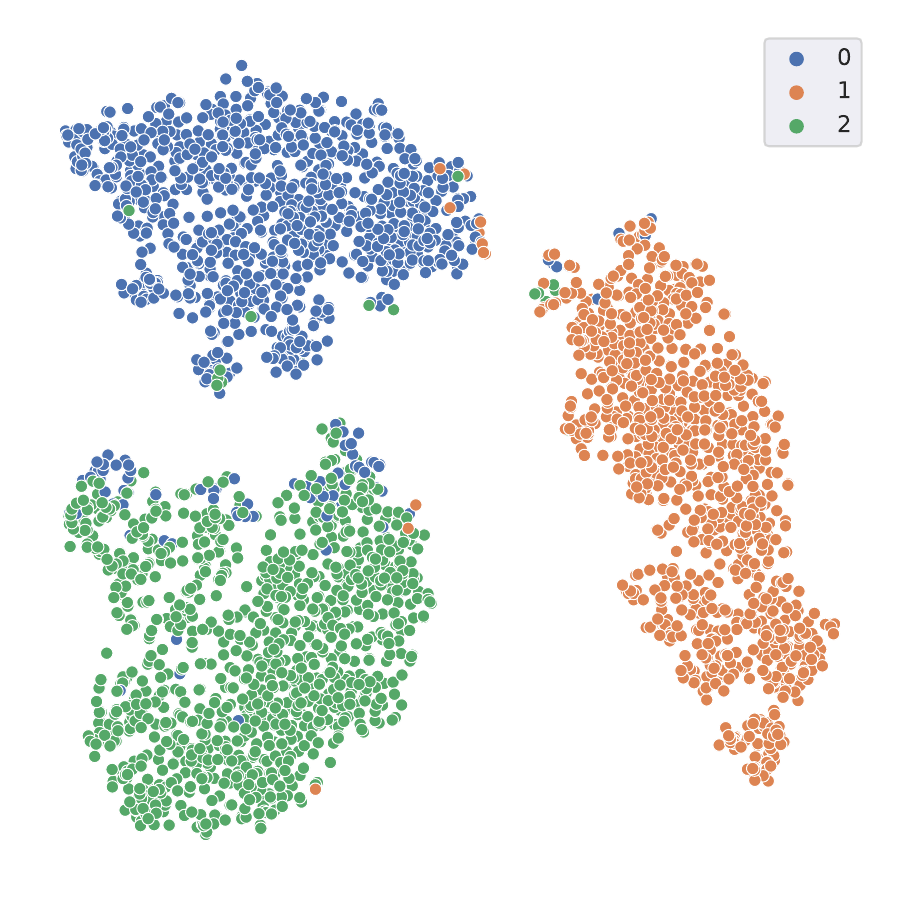}}
    \centerline{(b) Node embeddings}\medskip
  \end{minipage}
  \vspace{-0.4cm}
  \caption{Visualization of node embedding learned by our method on ACM dataset.}
  \label{fig:embedding}
    \vspace{-0.5cm}
  
\end{figure}

\begin{figure}[!t]
  \centering
  \begin{minipage}[h]{0.49\linewidth}
    \centering
    \centerline{\includegraphics[width=\linewidth]{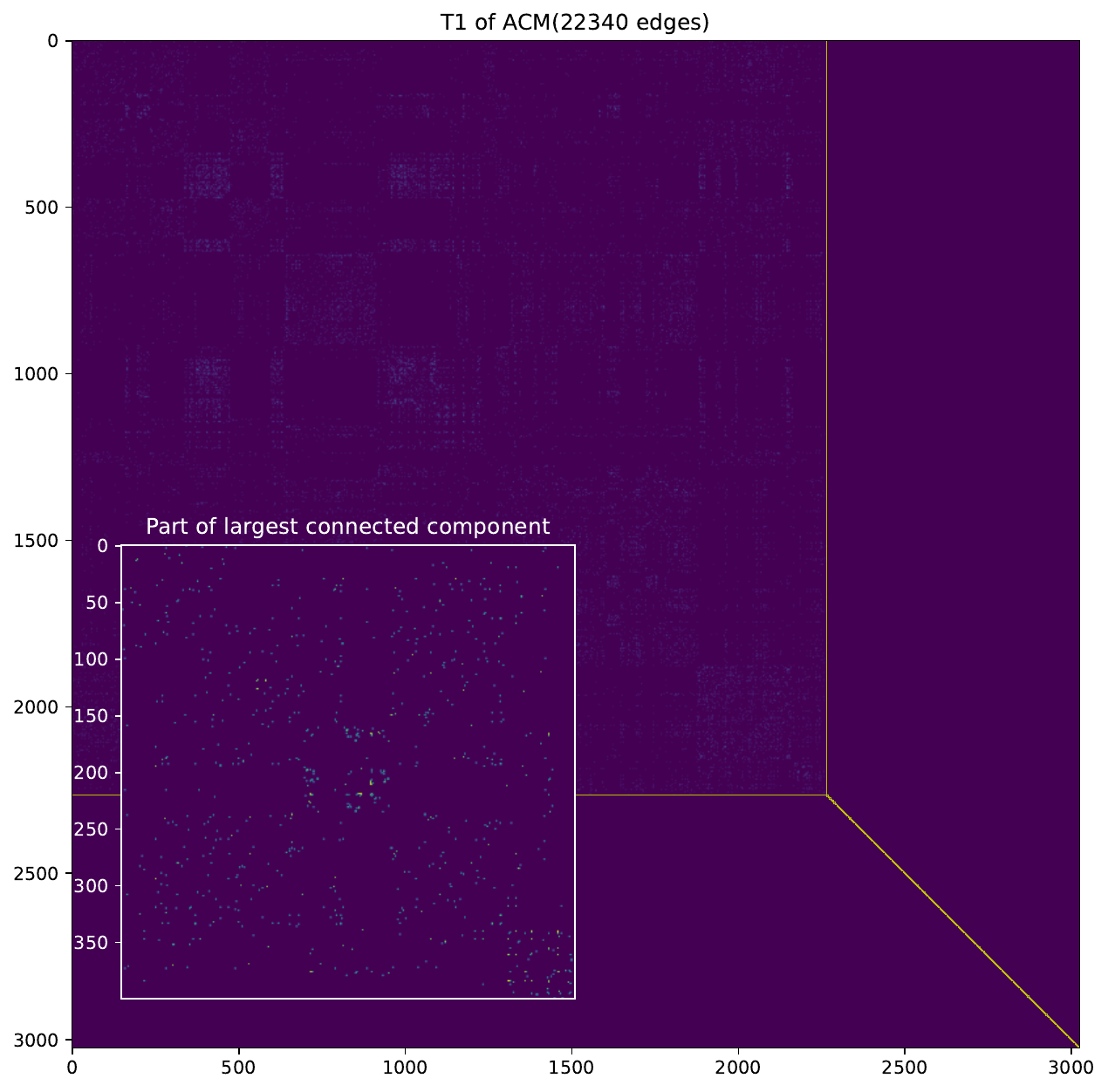}}
    \centerline{(a) $\mb{T}^{(1)}$}\medskip

  \end{minipage}
  \begin{minipage}[h]{0.49\linewidth}
    \centering
    \centerline{\includegraphics[width=\linewidth]{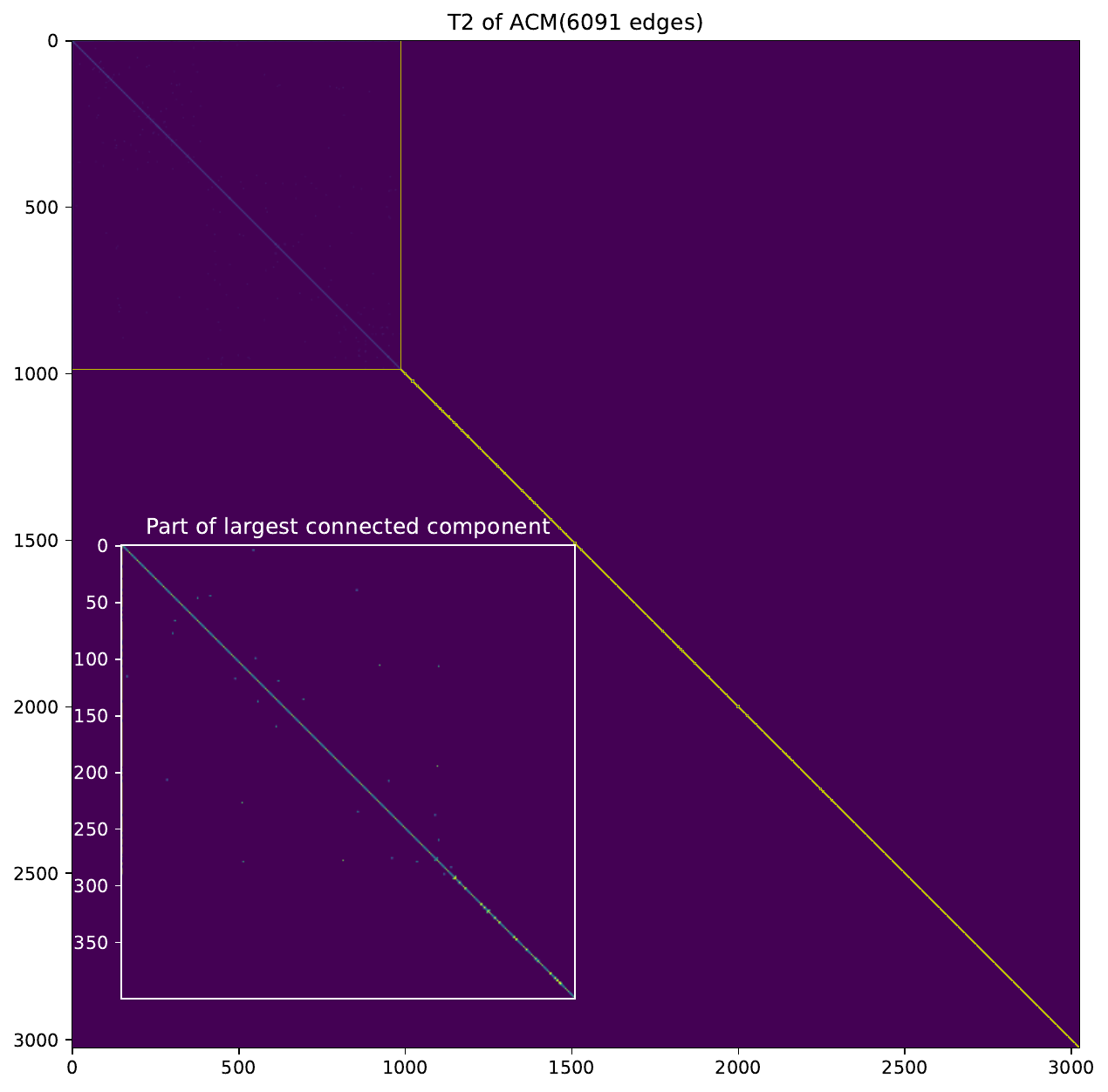}}
    \centerline{(b) $\mb{T}^{(2)}$}\medskip

  \end{minipage}
    \vspace{-0.4cm}
  \caption{Visualization of $\mb{T}^{(1)}$ and $\mb{T}^{(2)}$ learned by our method on ACM dataset. Yellow boxes represent nodes in the same connecting component.}
    \vspace{-0.5cm}
  \label{fig:topology}
  
\end{figure}

As Figure~\ref{fig:parameters} shows, our method has the same cross-view function as MIMO-GCN, but has much fewer parameters and much lower parameters' growth with the increase of K than MIMO-GCN. Our method also has comparable parameters amount with other simple methods such as P-GCN and M-GCN that have no cross-view function. Visualization of the learned node embedding is shown in Figure~\ref{fig:embedding}, which shown effective node embedding reflecting the node label information is learned by our method. Visualization of $\mb{T}^{(1)}$ and $\mb{T}^{(2)}$ learned by our method is shown in Figure~\ref{fig:topology}, which shown that $\mb{T}^{(1)}$ focus more on the global structures, while $\mb{T}^{(2)}$ focus more on the local structures.

\section{Conclusion}
In this paper, we propose a simple and efficient multigraph convolution method named as SMGCN. Different with previous works, SMGCN efficiently performs spatial cross-view message-passing via extracting credible cross-view topology, including edge-level and subgraph-level. It is worth noting that the proposed extraction methods for credible cross-view topology are just simple and effective instance. The idea, $i.e.$, extracting edge-level and subgraph-level to perform cross-view message-passing, has enlightening implications for the field of multigraph convolution fields. We really hope to see more effective extraction methods emerge in the future.

\bibliographystyle{ACM-Reference-Format}
\bibliography{sample-authordraft}
\end{document}